\documentclass[letterpaper, 10 pt, journal, twoside]{IEEEtran}
\usepackage{amsmath}
\usepackage{algorithmic}
\usepackage{algorithm}
\usepackage{array}
\usepackage[caption=false,font=footnotesize,labelfont=sf,textfont=sf]{subfig}
\usepackage{textcomp}
\usepackage{stfloats}
\usepackage{url}
\usepackage{verbatim}
\usepackage{graphicx}
\usepackage{cite}
\usepackage{tabularx}
\usepackage{graphicx}
\usepackage{ragged2e}
\usepackage{xcolor}
\usepackage{hyperref}
\usepackage{siunitx}

\newcommand{\orcc}{\includegraphics[height=\fontcharht\font`A]{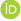}}
\newcommand{\orcid}[1]{\href{https://orcid.org/#1}{\orcc}}

\begin{document}

\title{A Neck Orthosis with Multi-Directional Variable Stiffness for Persons with Dropped Head Syndrome}

\author{Santiago Price Torrendell$^{1}$\orcid{0009-0002-4058-5958}, Hideki Kadone$^{2}$\orcid{0000-0003-2953-6434}, Modar Hassan$^{3}$\orcid{0000-0001-8466-6405}, Yang Chen$^{4}$\orcid{0000-0001-6204-9334},\\Kousei Miura$^{5}$\orcid{0000-0001-7826-184X}, and Kenji Suzuki$^{3}$ \orcid{0000-0003-1736-5404} 
\thanks{Manuscript received: February, 7, 2024; Revised March, 19, 2024; Accepted May, 5, 2024.}
\thanks{This paper was recommended for publication by Editor Yong-Lae Park upon evaluation of the Associate Editor and Reviewers’
comments.

This work was supported by the Japan Society for the Promotion of Science KAKENHI [grant number 23H00485], and the Ministry of Education, Culture, Sports, Science and Technology (MEXT) of Japan.} 
\thanks{$^{1}$Santiago Price Torrendell is with the Graduate School of Systems \& Information Engineering,
        University of Tsukuba, Tsukuba, Japan
        {\tt\small santiago@ai.iit.tsukuba.ac.jp}}%
\thanks{$^{2}$Hideki Kadone is with the Center for Innovative Medicine and Engineering, University of Tsukuba Hospital, Tsukuba, Japan
        {\tt\small kadone@ccr.tsukuba.ac.jp}}%
\thanks{$^{3}$Modar Hassan and Kenji Suzuki are with the Institute of Systems and Information Engineering, University of Tsukuba, Tsukuba, Japan
        {\tt\small modar@iit.tsukuba.ac.jp}, {\tt\small kenji@ieee.org}}       
\thanks{$^{4}$Yang Chen is with the Artificial Intelligence Laboratory, University of Tsukuba, Tsukuba, Japan
        {\tt\small chenyang@ai.iit.tsukuba.ac.jp}}  
\thanks{$^{5}$Kousei Miura is with the Department of Orthopaedic Surgery, Faculty of Medicine, University of Tsukuba, Tsukuba, Japan
        {\tt\small kmiura@tsukuba-seikei.jp}}%
\thanks{Digital Object Identifier (DOI): see top of this page.}
}

\markboth{IEEE Robotics and Automation Letters. Preprint Version. Accepted May, 2024}%
{Price Torrendell \MakeLowercase{\textit{et al.}}: A Neck Orthosis with Multi-Directional Variable Stiffness for Persons with Dropped Head Syndrome}

\IEEEpubid{\begin{tabular}[t]{@{}l@{}}2377-3766 ~\copyright~2024 IEEE. Personal use of this material is permitted. Permission from IEEE must be obtained for all other uses,\\in any current or future media, including reprinting/republishing this material for advertising or promotional purposes, creating new\\ collective works, for resale or redistribution to servers or lists, or reuse of any copyrighted component of this work in other works.\end{tabular}}


\maketitle

\begin{abstract}
Dropped Head Syndrome (DHS) causes a passively correctable neck deformation. Currently, there is no wearable orthopedic neck brace to fulfill the needs of persons suffering from DHS. Related works have made progress in this area by creating mobile neck braces that provide head support to mitigate deformation while permitting neck mobility, which enhances user-perceived comfort and quality of life. Specifically, passive designs show great potential for fully functional devices in the short term due to their inherent simplicity and compactness, although achieving suitable support presents some challenges. This work introduces a novel compliant mechanism that provides non-restrictive adjustable support for the neck's anterior and posterior flexion movements while enabling its unconstrained free rotation. The results from the experiments on non-affected persons suggest that the device provides the proposed adjustable support that unloads the muscle groups involved in supporting the head without overloading the antagonist muscle groups. Simultaneously, it was verified that the free rotation is achieved regardless of the stiffness configuration of the device.
\end{abstract}


\section{INTRODUCTION}
\subsection{Background}
\IEEEPARstart{D}{ropped} Head Syndrome (DHS) occurs in a set of neurological, muscular, or neuro-muscular conditions affecting the paraspinal muscles. This syndrome alters the normal neck posture where the head is shifted forward, technically known as cervical kyphosis, which leads to chin-on-chest deformity in extreme cases.

This uncommon inclination of the head has multiple negative effects on the affected persons' daily life by causing horizontal gaze disorders, gait imbalance, and even dysphagia and dyspnea for the more severe cases \cite{Igawa_Ishii_Isogai_Suzuki_Ishizaka_Funao_2021}. Traditionally, the syndrome has been initially addressed by providing a passive correction using a neck brace \cite{Brodell2020}, which excessively constrains the cervical range of motion. This restriction creates discomfort for the user eventually rejecting the device \cite{Spears}.

Surgical intervention is a viable alternative when the non-invasive approach is ineffective. This practice shows general long-term improvements but it also often involves post-surgery complications \cite{Brodell2020}. In this context, there has been an increasing interest in assistive head orthosis that provides mobile support which could increase the efficacy of the non-invasive approach \cite{AROCKIADOSS2023101306}. Moreover, these new generation braces brought a paradigm change for solutions that simultaneously mitigate neck deformation and restore lost mobility related to neck muscle weakness. 
\IEEEpubidadjcol
\begin{figure}[t]
    \begin{center}
    \includegraphics[width=1\linewidth]{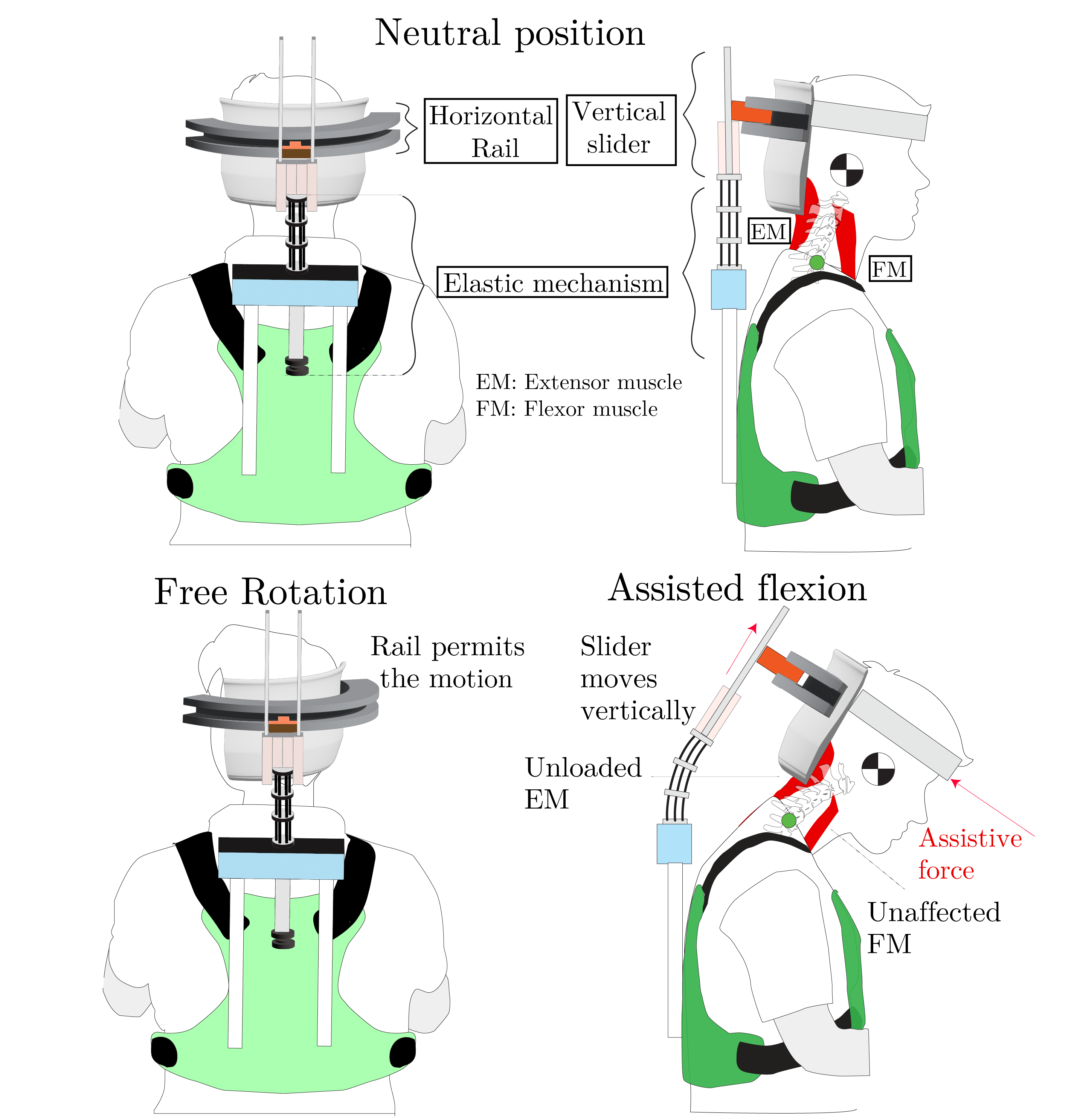} %
    \caption{Illustration of neck mobility with the proposed elastic mechanism and components of the exoskeleton device. The elastic actuator provides a nonrestrictive force that partially compensates the head weight during sagittal flexion relieving the extensor muscles. The assistive force is regulated by adjusting the stiffness of the elastic mechanism.}
    \label{fig:head_assitive_eff}
    \end{center}      
    \vspace{-0.6cm}
\end{figure} 
\subsection{Related Works}
In recent years, several promising assistive braces have been proposed with passively actuated designs that are simpler and more robust solutions than active systems 
\cite{AROCKIADOSS2023101306}.

One of the most prominent works in the field is the Sheffield Snood brace, the only flexible brace certified as a medical device class 1 \cite{Pancani2016}. The design consists of a snood made of flexible fabric with removable beam-like structures which provide a customized elastic support that permits flexion and rotational motions of the head \cite{Pancani2016}. A cross-sectional study on 139 neck brace users showed a clear preference for the flexible snood over the brace used before \cite{Sproson2021}. However, the results from this questionnaire showed that the brace's effects on breathing and non-swallowing are non-negligible which has been related to the contact of the anterior side of the neck with the brace by other studies \cite{Spears}.

Two other prominent works proposed an alternative approach from the Sheffield brace by compensating the head weight in multiple positions using a rigid link mechanism attached to the user's head through a headband that does not require chin support.

Mohammed et al. developed a passive wheelchair device for DHS patients that provides support on the sagittal plane for a wide range of motion while enabling free neck rotation \cite{Mahmood2021}. Electromyographic studies on healthy participants showed that wearing the device caused a significant reduction and increment in the average activity of the extensor and flexor muscles respectively for holding the head in different postures.

Haohan et al. introduced a lightweight three-DoF parallel mechanism to assist the head motion of patients suffering from Amyotrophic Lateral Sclerosis. The transmission design optimizes the device's DoF to the head workspace resulting in achieving 70\% of the range of head rotations \cite{Zhang2017}. The original mechanism was further developed as an active and passive device. The active version employs three motors controlled through a joystick that achieved muscle effort reduction in healthy participants in preclinical trials \cite{Zhang2018} and also enhances the postural control of end users \cite{Zhang2022}. Additionally, the passive version utilizes torsional springs that provide elastic support free from control interfaces but its effects on muscle effort reduction were not statistically significant \cite{Zhang20183}. 
\subsection{Contribution}
The presented works show promising results in this new field of research and highlight the feasibility of a functional passive assistive brace for DHS patients. However, no prior studies have presented strong evidence that the proposed device provides suitable support that unloads the weak muscles without overloading the antagonist group which may exacerbate muscle unbalance of the neck and aggravate the neck deformity. The optimal mechanisms to support the neck muscles, and the amount of neck mobility that can be preserved while supporting DHS are still unknown. In this work, we propose and investigate a neck exoskeleton that provides support through a compliant structure based on flexible rods to address the presented research gaps.

This research aims to develop a compliant mechanism that provides the critical support needed for DHS while still recruiting the weak muscle groups, and preserving neck mobility as much as possible. This work presents the mechanical design of the system, the physical model to estimate the required moment to assist the head, a simple formulation of the mechanical behavior of the passive mechanism along with the experimental validation, and non-impaired participants study where the effects of the actuator stiffness on the EMG activity of several neck muscles were studied.

\section{METHODS}
\subsection{Mechanical design}
The device requirements were defined through interviews with a physical therapist experienced in DHS, resulting in a brace with two distinct behaviors. On the Sagittal plane, the device offers elastic support to correct neck deformation, allowing users to maintain a straight head position. This adjustable support should accommodate each patient's condition without exerting a compressive force on the spine, which could hinder recovery. Horizontally, the mechanism allows free neck rotation for user comfort, which is safe for the person.

At this research stage, we prioritized a robust support structure over a lightweight and compact design to ensure accurate actuator assessment, avoiding potential issues like elastic deformation and loose 
fittings that could compromise device performance. Another core design aspect was using adjustable unions that minimize the fitting time of the device.
\begin{figure}[h]
    \begin{center}
    \includegraphics[width=0.95\linewidth]{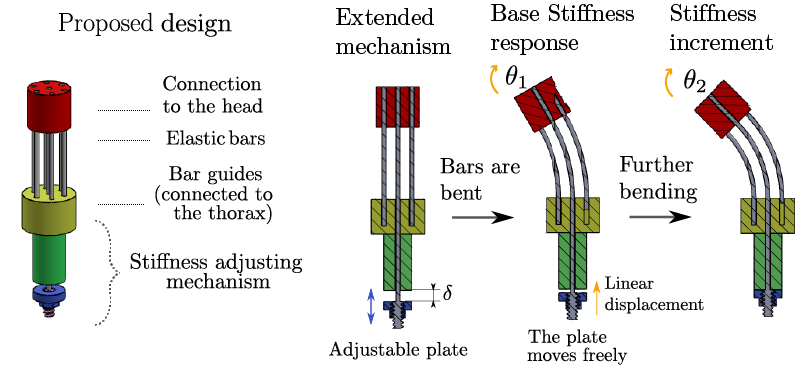} %
    \caption{Schematic view of the passive actuator and working principle in three sectioned views}
    \label{fig:act_sch}
    \end{center}      
    \vspace{-0.5cm}
\end{figure} 

The developed device is presented in Figure \ref{fig:head_assitive_eff} which illustrates the behavior when the user moves the head. On the Sagittal plane, the elastic mechanism bends during the neck flexion providing an assistive force that supports the motion while partially compensating the head weight. This assistance would relieve the extensor muscles that hold the head without overloading the flexor muscles that perform the motion. Due to the eccentricity between the mechanism and the cervical vertebrae, a vertical rail connects the mechanism to the head attachment to minimize both restricting the head motion and compressing the spine. This element is linked to the horizontal rail of the head attachment and decouples the head from the system during the neck rotations. Between both sliders, a universal joint prevents the mechanism from transmitting undesired torques to the head.

Regarding the elastic mechanism actuation, the system has two well-differentiated elastic behaviors where the flexible stiffness of the bar array is controlled by restricting the ascendant movement of the central bar's lower end as shown in the diagram from Figure \ref{fig:act_sch}. The vertical and bending motions are tightly coupled \cite{Me10122051}. The sliding mechanism transmits normal forces to the bars deflecting them like a cantilever beam. For a low bending angle $\theta_1$, the stiffness is relatively low as the central bar can move freely, but when the component touches the cylinder at the angle $\theta_2$ the vertical motion is restricted leading to a sudden stiffness increase. The bending angle where the mode transition occurs will depend on the initial gap between the cylinder and the plate which is $\delta$. The smaller the value of $\delta$ is, the earlier the elastic transition will occur.

\subsection{Modeling}
\subsubsection{Estimation of the assistive moment}
\begin{figure}[t]
    \begin{center}
     \includegraphics[width=0.95\linewidth]{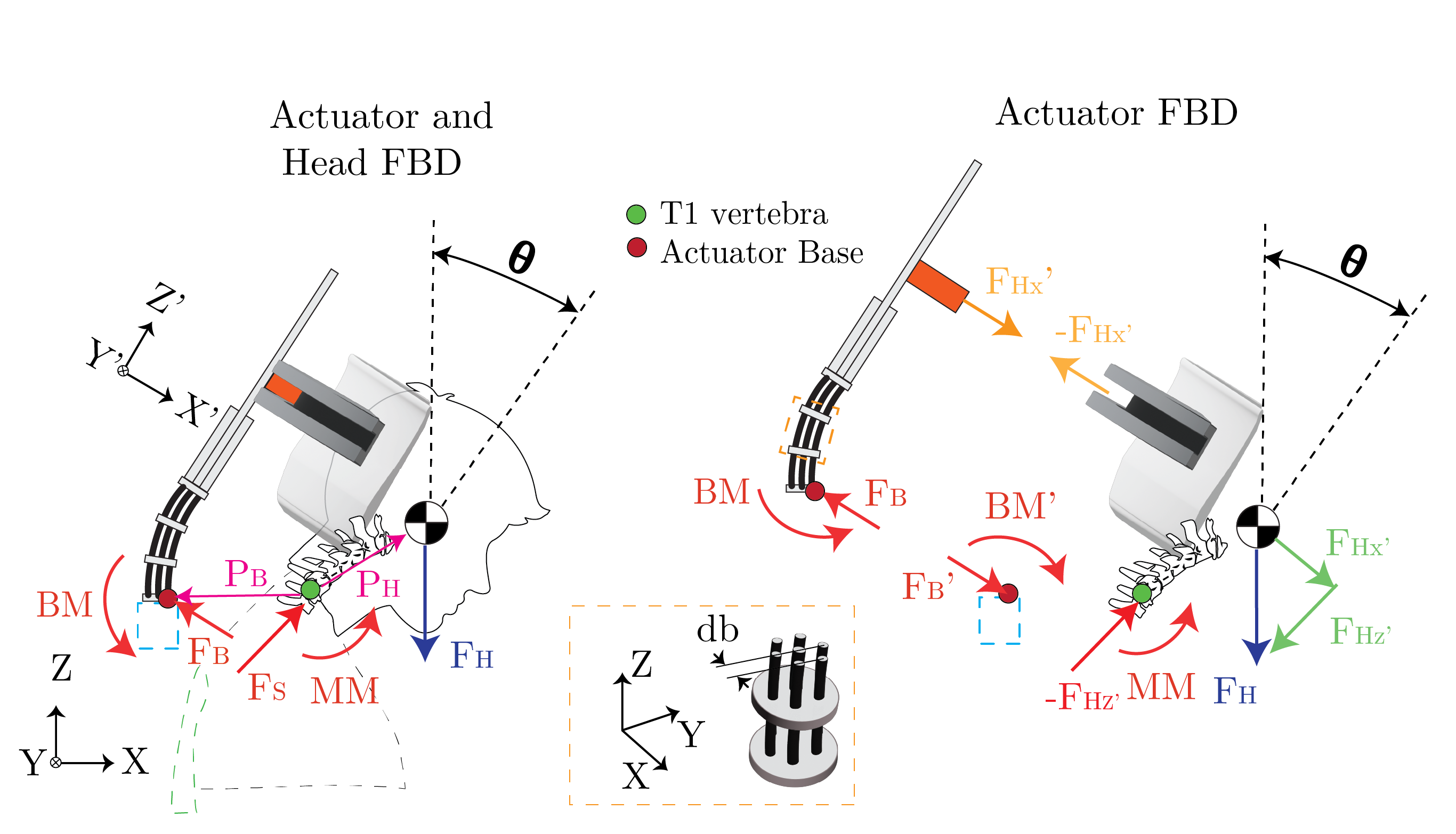}
         \caption{Free Body Diagram showing the forces involved in the interaction between the device and the user.}
         \label{fig:FBD}
    \end{center}      
    \vspace{-0.6cm}
\end{figure}

As the main purpose of the elastic mechanism is to assist the head motion on the sagittal plane, the required assistive moment should ideally balance the head weight during such a motion. The moment`s magnitude depends on the head inclination about the body frame and is estimated through the free body diagrams (FBDs) of Figure \ref{fig:FBD}. The Actuator and Head FBD shows that the assistive moment should counterbalance the torques from the reaction force at the base of the actuator $F_B$ and the head weight $F_H$, which is quantified in a moment balance about the C7-T1 level of the cervical spine, based on previous work \cite{Vasavada1904}, on the direction of the Y axis as shown in the expression \ref{Eq:Mom_bal}:
\begin{align}
 \begin{split}
     \sum My_{C7}=0 \rightarrow \text{MM}&=\vec{F_H}\times\vec{P_H}-AM\\
     AM&=BM-\vec{F_B}\times\vec{P_B} 
 \end{split}
\label{Eq:Mom_bal}
  \end{align}
  This balance shows the muscle moment MM counteracts the torque due to the head weight, mitigated by the assistive moment AM. AM results from the difference between the base moment BM and the eccentric base force $F_B$ to the vertebral joint. Under this formulation, the device's purpose is to minimize MM to relieve the effort of the extensor muscles. A device`s stiffness superior to the ideal value would overload the flexor muscles to exert a negative MM to keep the head tilted.
  The direction and magnitude of $F_B$ are deduced from the Actuator diagram. As the elastic structure of the bars keeps the vertical guide aligned to the head vertical axis Z', $F_B$ counter-balance the horizontal component of the head weight $F_{Hx'}$ which is transmitted to the actuator basement. On the perpendicular axis Z',  the spine compensates the component $F_{Hz'}$ while the slider does not transmit compression effort.
  As a result, the system is in equilibrium as the forces on the Z' and X' direction is zero as the expression \ref{Eq:F_bal} shows.
 \begin{align}
 \begin{split}
     \sum F_{X'}&=0 \rightarrow \vec{F_B}+\vec{F_{H_{X'}}}=0 \\
     \sum F_{Z'}&=0 \rightarrow \vec{F_S}+\vec{F_{H_{Z'}}}=0
 \end{split}
\label{Eq:F_bal}
  \end{align}  
Based on this model, the ideal moment to fully compensate for the head weight was estimated from head motion measurements as explained in the experiment section.

\begin{figure}[t]
    \begin{center}   \includegraphics[width=0.2\textwidth]{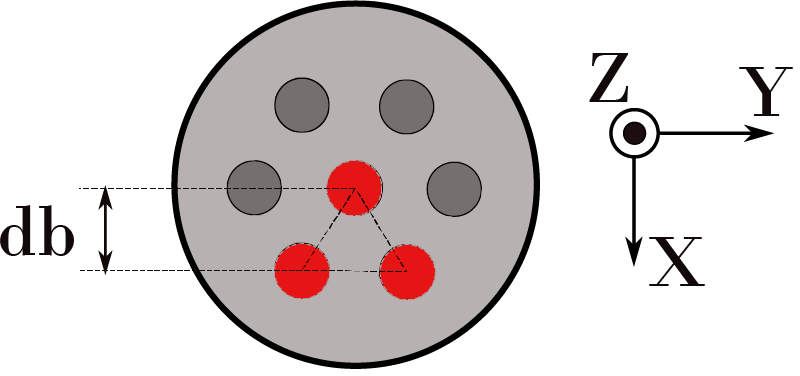}
         \caption{Cross section of the 7 actuator`s bars in a hexagonal arrangement. The 3 red bars are coupled during the High-Stiffness mode.}
         \label{fig:bars_CS}
    \end{center}      
    \vspace{-0.6cm}
\end{figure} 

\subsubsection{Mathematical model of the actuator}
The model from the original work was adapted to the new design to relate the system stiffness with the material properties and geometric parameters. 
Under the hypothesis that the separation of the bars remains constant during bending due to the spacers, the Belendez solution \cite{Belen2002} for a single bar deflection was used to predict the device response with expression \ref{Eq:sbar_equation}. This approach omits the bars length variation (1-4 mm) during flexion.\\
 \begin{align}
 \begin{split}
     M_{B}(\theta) & = \Gamma(\theta)\cdot E \cdot I_{eq} \cdot \sqrt{\sin(\theta)} / L\\
      \Gamma(\theta) &= \int_0^{\theta}\dfrac{d\gamma}{\sin(\theta)-\sin(\gamma)}
 \end{split}
\label{Eq:sbar_equation} ,
  \end{align}
where $\theta$ and L are the bending angle and length of the bars, respectively. $I_{eq}$ is the equivalent moment of inertia that depends on the device`s stiffness condition. $I_{eq}$  is defined for two stiffness states used during the device experimental validation: the Loaded and Base Stiffness.

During the Base Stiffness Mode (BS), the vertical motion of the central bar is unrestricted regardless of the bending angle $\theta$, all the bars work in parallel presenting a similar curvature.  Each bar contribution to the assistive torque is equal and is equivalent to the response of an ideal bar $M_{B}^{BS}$ whose second moment is the sum of each real bar contribution as the expression \ref{eq:lowS_I} shows.
 \begin{align}
 \begin{split}
    M_{B}^{BS}(\theta)\rightarrow I_{eq}=\sum ^{n_{\text{b}}}_{i=1}I_{i} \\
\end{split}
    \label{eq:lowS_I},
\end{align}
where $I_{i}$ is the moment of inertia of each beam's cross-section about its neutral axis \cite{Belen2002} and $n_b$ is the total amount of bar which is 7.

In the Loaded Stiffness Mode (LS), the separation $\delta$ is 0 so the central bar bottom is completely constrained to move, setting a stretching effort along the bar. In reaction to the pulling force, the bars on the concave side, left side in Figure \ref{fig:act_sch}, will experience a compressive effort to balance the inner reaction on the system. Assuming that the coplanar cross sections of the three coupled bars remain coplanar during flexion, they can be treated as part of a single elastic beam whose cross sections are the three circular bar sections arranged in a triangular pattern as shown in Figure \ref{fig:bars_CS}.

The second moment of the triad bars is determined by the Steiner theorem which combined with the other three decoupled bars defines the $I_{eq}$ when exerting the torque $M_{B}^{LS}$ as shown in the expression \ref{Eq:highS_I}.
 \begin{align}
 \begin{split}
     M_{B}^{LS}(\theta)\rightarrow I_{eq} & = \sum ^{n_b-n_c}_{i=1}I_{i} + I_{3bs} \\
     I_{3bs}&=D_{b}^2 \cdot S_{b} \cdot 2 / 3 + \sum ^{n_{c}=3}_{i=1}I_{j} 
 \end{split}
\label{Eq:highS_I},   
  \end{align}
where $D_{b}$ is the separation of the bars on the ZX plane shown in Figure \ref{fig:FBD}, $S_{b}$ is the area of every single bar, and $n_{c}$ is the number of coupled bars. 

The estimations for $M_{B}^{BS}$ and $M_{B}^{LS}$ are contrasted to experimental data in Figure \ref{fig:mech_char_res}.

\subsubsection{Device`s prototype}
The actuator prototype features 1.5mm diameter carbon fiber bars (Unxell) with an 80mm free length. Other components were 3D-printed in polycarbonate (Fortus 350, Stratasys) and assembled using epoxy adhesive ( J-B KWIKWELD). The device's performance was evaluated in a bending test, as described below.

\begin{figure}[t]
    \begin{center}
     \includegraphics[width=0.9\linewidth]{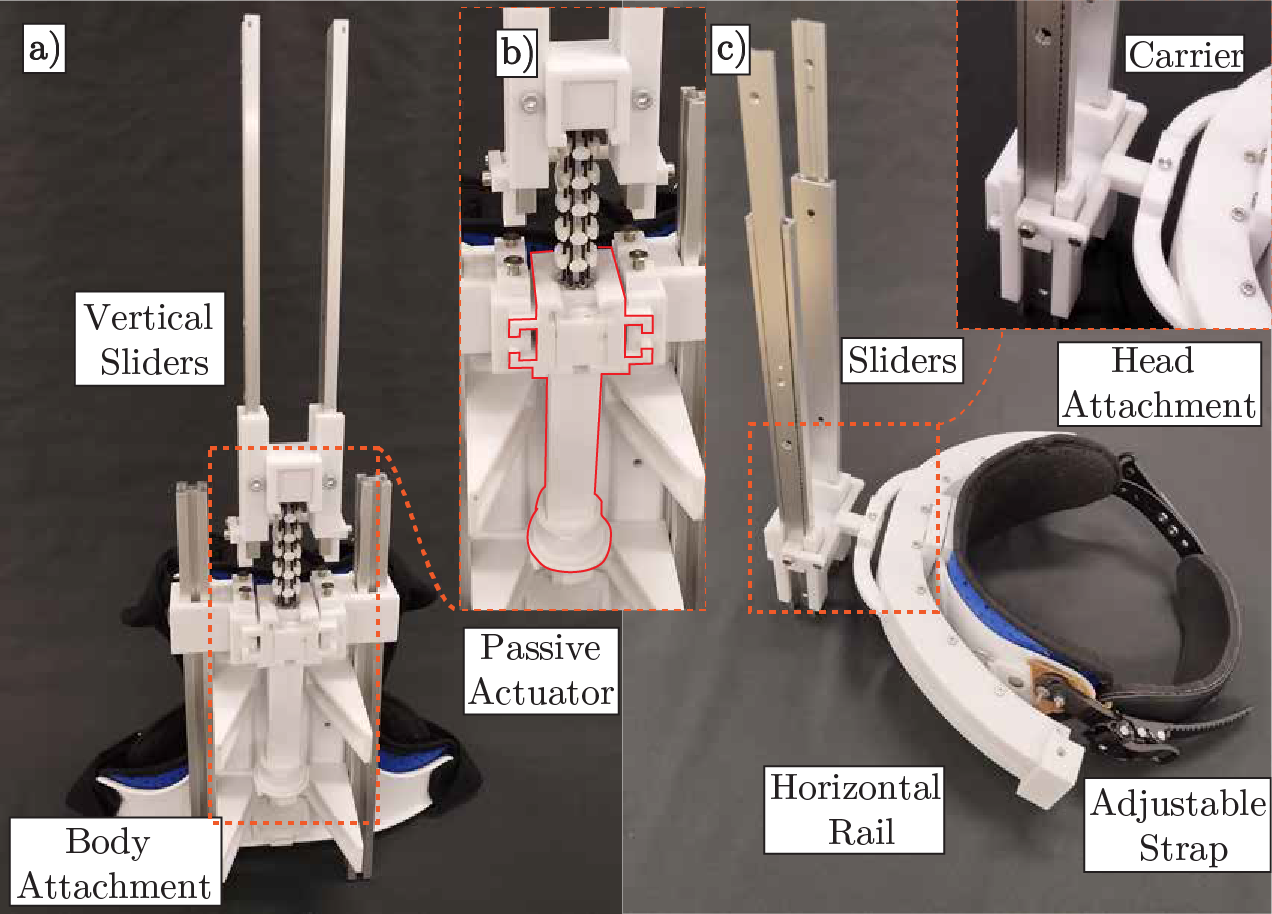}
         \caption{Neck brace components. (a)Body attachment with the passive actuator and slider b) Isolated actuator constituted by the elastic bars inserted in the body attachment. The middle bar connects to the mobile plate at the bottom. c) Head attachment and horizontal rail connected to the vertical slider.}
         \label{fig:dev_exp}
    \end{center}      
    \vspace{-0.9cm}
\end{figure}

The device used in the user`s experiments consists of a commercial rigid halo neck brace (Brand: Hengshui Jingkang, model N211, Instrument classification: Class I) that incorporates the actuator and transmission mechanism as shown in Figure \ref{fig:dev_exp}. The mounting and adjustment can done by a caregiver in under 4 minutes. The body attachments accommodate users with a xiphoid circumference of 70-100 cm and a head circumference of 70-80 cm, reflecting the average body dimensions of the Asian elderly population \cite{HU2007303}, who are potential users of this device. 

\section{EXPERIMENTS}
\subsection{Estimation of the actuator's ideal moment:}
 Based on the expression \ref{Eq:Mom_bal}, the ideal moment at the base of the actuator BM  to balance the head weight without muscle effort (MM=0) was estimated from a single-participant experiment. The protocol involves measuring the head motion using a motion capture system while performing continuing flexion-extension cycles for one minute. The markers were placed on the head and body based on previous research \cite{Zhang2017}.
 
 
 Under this setup, the head inclination angle $\theta$ is measured, which defines the X' component of the head weight $F_{H_{X'}}$ that is the opposite of the $F_B$ according to \ref{Eq:F_bal}.
 
 After finishing the trial, the relative position of the markers of the actuator base and the center of mass of the head were elucidated from a side-view picture of the participant`s head while wearing the markers. Therefore, $P_B$ and $P_H$ are computed from the markers' data. 

 The relative location of the actuator base was estimated at the intersection of a horizontal axis passing through the c7 vertebra and a vertical one tangent to the scapula contour. 
 On the sagittal plane, the head center of mass was referred from the Frankfort and Beuviex Planes as outlined in \cite{bussone2005linear}. 
 By estimating the head gravitational force $F_H$ to 50N, the ideal moment $BM_{id}$ is computed using the expression \ref{Eq:Mom_bal}.
   \begin{align}
 \begin{split}
          BM_{id}&=\vec{F_H}\times\vec{P_H}-\vec{F_B}\times\vec{P_B} 
\end{split}
    \label{eq:ideal_M},
\end{align}
 This prediction was contrasted with the mechanical characterization results in Figure \ref{fig:mech_char_res}.

\begin{figure}[t]
    \begin{center}
     \includegraphics[width=0.9\linewidth]{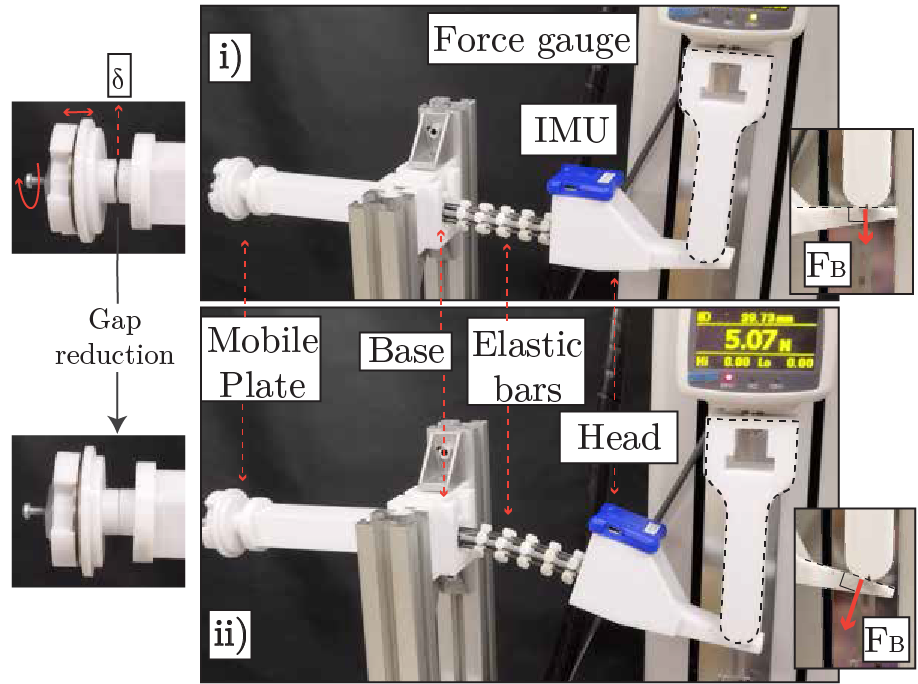}
         \caption{Prototype tested during the mechanical characterization. The test machine applies a controlled load on the head to bend the bars. Under this action, the gap cavity reduces until the mobile plate contacts the upper body, which stiffens the actuator. The bending angle where the stiffening happens is defined by $\delta$, which is adjusted by rotating the plate.}
         \label{fig:bent_proto}
    \end{center}      
    \vspace{-0.6cm}
\end{figure} 
\subsection{Mechanical characterization}
The prototype's mechanical rigidity was characterized by conducting a bending test across various increments of the distance parameter ($\delta$). In this assessment, controlled loads were applied perpendicular to the upper attachment of the elastic mechanism shown in Figure \ref{fig:bent_proto}, mimicking the standard functionality of the entire system. These loads were provided by a conventional motorized test stand by IMADA, equipped with a force gauge and a displacement sensor (IMADA MX2-2500N-FA, ZTA-50N). The evaluation involved determining the moment around the actuator base. Notably, the base was also used as a reference to estimate the assistive moment in the static analysis. Lever calculations were derived from the displacement data recorded by the test stand and the bending angle of the bars, the latter being measured using an IMU (LP-research LPMS-B2).

\begin{figure} 
    \centering
  \subfloat[\label{fig:ex_set} Experimental setup]{%
       \includegraphics[width=0.44\linewidth]{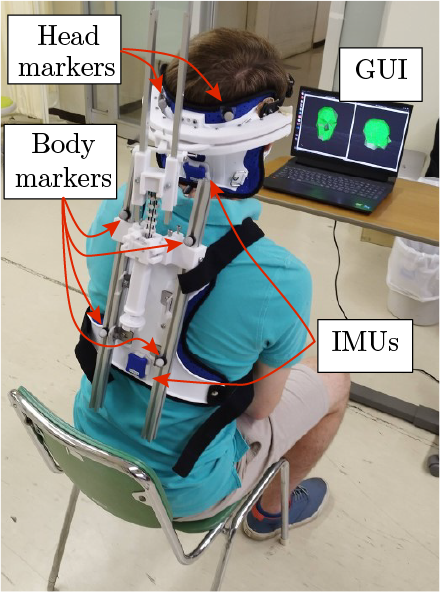}}
    \hfill
  \subfloat[\label{fig:EMG_placement} EMG sensors placement]{%
        \includegraphics[width=0.45\linewidth]{Fig/exp_setup_2.png}}
  \caption{User experiment. The participants do a series of randomized head motions guided by the user interface. The IMUs drive the interface and The MOCAP markers and the EMG provides synchronized data of the head motion and the EMG activity of the muscles.}
  \label{fig:exp_setup}
\end{figure}

\subsection{Human Experiment} 
Under the recommendation of a doctor who specialized in DHS, we did a Biomechanical evaluation of the device as a prior step to a clinical study that involves persons with DHS. The experiment, which involved 8 volunteers with no history of neck disorders (7 males and 1 female, age 25.4$\pm$3.2 years, 173.6$\pm$7.3 cm, 68.8$\pm$12.5 Kg), was conducted to evaluate the brace performance on reducing muscular effort in the neck while performing head motions that were guided by a Graphical User Interface (GUI). This experiment was approved by the Research Ethics Board of the University of Tsukuba. (2022R680) 

This experiment compares two stiffness conditions to solely analyze the adjustable support provided by the elastic mechanism. This approach minimizes the effects of the mechanism imperfection in the comparison given that this factor similarly affects both conditions. For the baseline, the mechanism was set to the Base Stiffness mode which is the least rigid state of the device. Secondly, the device was set to the Loaded Stiffness mode.  

The purpose of the evaluation was to determine the device's efficacy in assisting the head during flexion on the sagittal plane while permitting the free rotation of the neck. This behavior was studied based on three hypotheses: (a) During Sagittal flexion, healthy participants experience an EMG reduction on the muscles supporting the head when increasing the brace stiffness. (b) During axial rotation, the users do not experience significant changes in muscle activity regardless of the device condition. (c) The device`s stiffness does not significantly affect neck mobility. 

\begin{figure}[t]
    \begin{center}
    \includegraphics[width=1\linewidth]{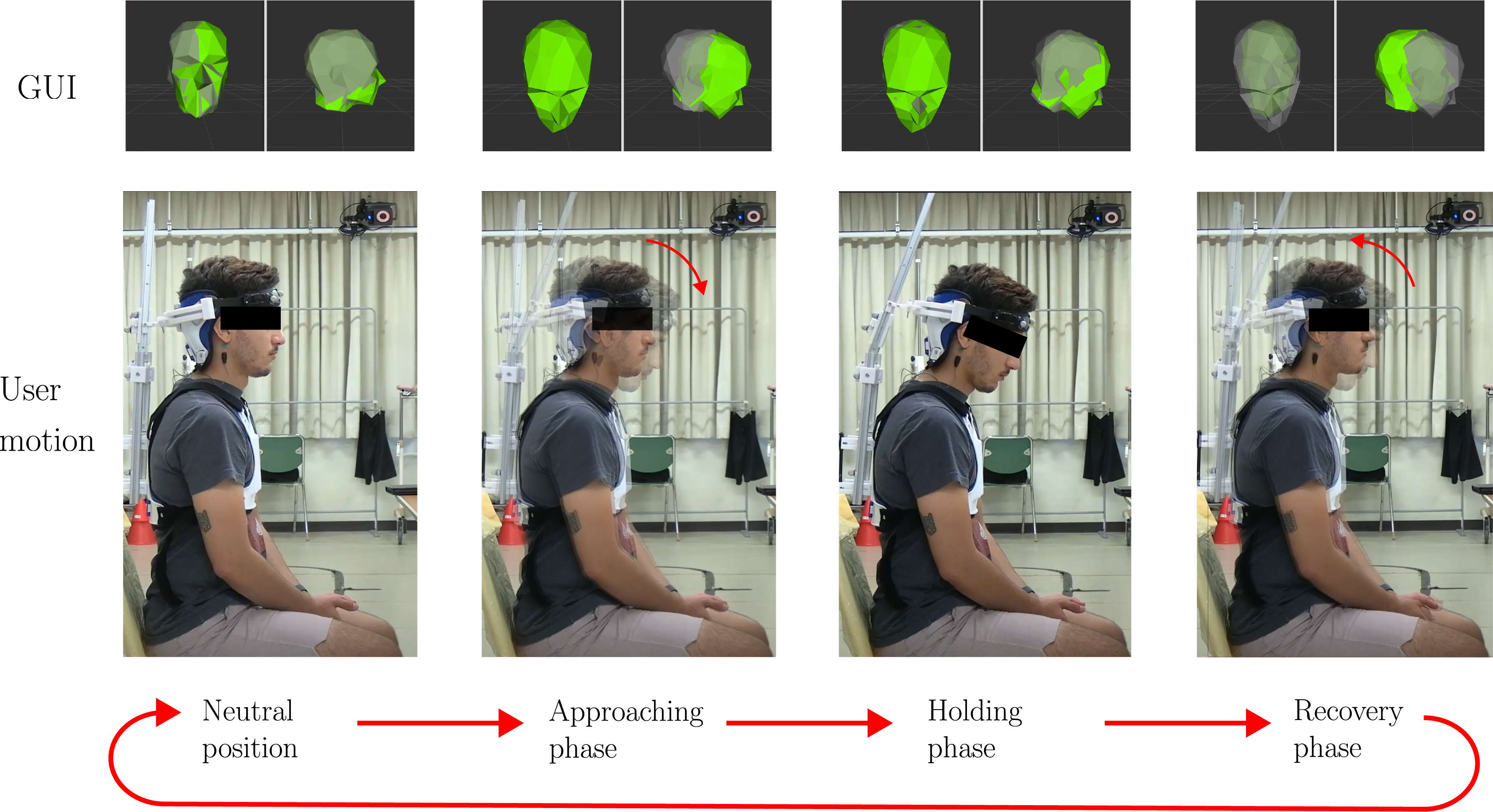} %
        \caption{Motion guidance provided by the interface. The user controls the position of the grey marker and the green guides the movements. During a cycle, the reference marker moves, and the user has to match both avatars (Approaching phase). When the target is reached, the user keeps the head position for 10 seconds (Holding phase). Finally, the reference  marker returns to the neutral position and the user follows (Recovery phase)}
    \label{fig:GUI_seq}
    \end{center}      
    \vspace{-0.6cm}
\end{figure} 

\subsubsection{Instrumentation}
The muscle EMG activity was measured using Trigno Mini surface electrodes (20–500 Hz, Common Mode Rejection Ratio of 80 dB, input impedance exceeding 1000 $\Omega$, Delsys Inc., Boston, MA). After properly cleaning the skin in contact with the electrodes, the sensors were mounted on both Stercleidomastoid (SCM) and Splenius Capitis (SPL), which are the main superficial muscles involved in head mobility \cite{Yajima2022}. As Figure \ref{fig:EMG_placement} shows, the acquisition electrodes for each SCM muscle were placed along a line drawn from the sternal notch to the mastoid process, at 1/3 the length of the line from the mastoid process \cite{SOMMERICH2000377}. The electrodes that measure the SPLs were placed on the palpable portion between the upper trapezius and SCM located between 6 and 8 cm lateral of the median line at the level of C4 \cite{SOMMERICH2000377}.

The head kinematic was measured by adding motion capture markers (MX System, Vicon Motion Systems, Ltd., U.K.) on the body and head attachment as shown in Figure \ref{fig:exp_setup}.

Simultaneously, the GUI received current head orientation about the body in real-time by one pair of IMU sensors (LPMS-B2 Life Performance Research, Tokyo, Japan) placed on both attachments, as Figure \ref{fig:ex_set} shows. 

The visual interface is based on the package RVIZ from ROS (Noetic Nynjemy distribution). The IMU data acquisition was computed using the library OpenZen\footnote{\url{https://bitbucket.org/lpresearch/openzenros/src/master/}}.
\subsubsection{Protocol}
The experimental procedure is mainly based on the work of M.N. Mahmood et al, \cite{Mahmood2021} and comprises two sessions of 5 minutes that evaluate the device performance when using the device in two different configurations. The device was set to the Base Stiffness mode in the first session and the Loaded Stiffness in the second Session. A break time of 5 minutes between sessions was set to prevent the participants' muscular fatigue.

During each trial, the participant moves the head following the instructions of the interface on the monitor in front of them. As a first step, there is a 10-minute practice session to familiarize the participant with the system while wearing the device. The interface presents the frontal and side view of two markers with the shape of a human head. The gray marker mimics the user`s head orientation measured by the IMUs while the green guides the prescribed motion. The participant's task is to follow this reference and make both markers match. Figure \ref{fig:GUI_seq} illustrates the actions for completing one motion cycle from the sequence, which can be divided into three phases: approaching, holding, and recovery. Initially, the user keeps the head straight and both markers' positions stay neutral. After 3 seconds, the approaching phase begins when the green marker orientation changes and the user moves the head intending to match both markers again. The marker matching initiates the holding phase where the user`s marker is kept near the target for 10 seconds. After the elapsed holding time, the recovery phase starts where the reference returns to the neutral position, and the user follows.\\    
The reference marker repeats the described cycle in a randomly ordered sequence involving twelve postures during the holding phase, each repeated twice. The included head inclinations were $\pm$15 and $\pm$40 degrees on the Sagital plane that corresponds to frontal and back flexion, $\pm$15 and $\pm$30$^{\circ}$ on the Transverse plane for right and left rotation,(15 and 40$^{\circ}$) on the Coronal plane. \\
The user performed these movements guided by the GUI as indicated in Figure \ref{fig:GUI_seq} to keep the head at each position for 10 seconds.

\subsubsection{Postprocessing}
\begin{figure}[t]
    \begin{center}
    \includegraphics[width=1\linewidth]{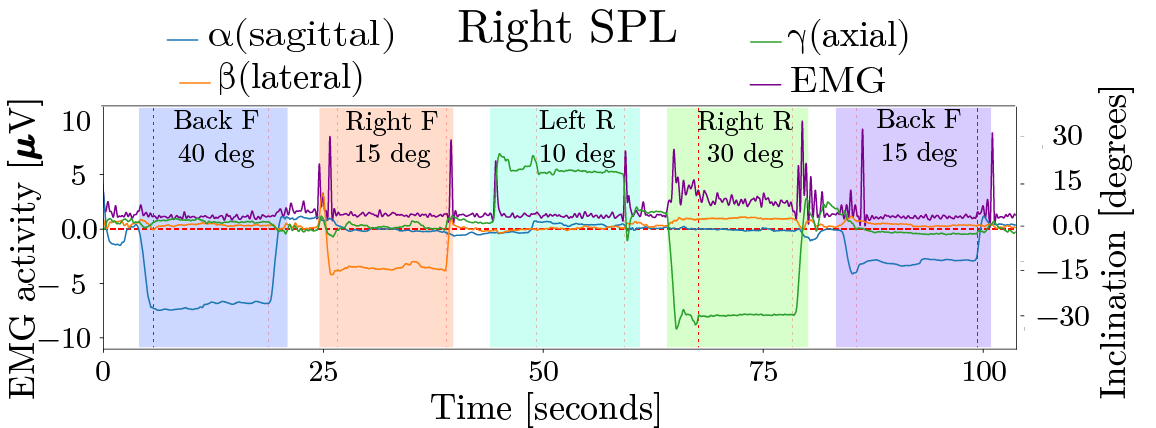} %
    \caption{Segmented head orientation and EMG activity during Flexion (F) and Rotation (R). The three phases are delimited by vertical dashed lines in each colored motion cycle.} 
    \label{fig:MOCAP_EEMG_segm}
    \end{center}      
    \vspace{-0.6cm}
\end{figure}

After the experiment is concluded, the EMG data is manually segmented using the head orientation calculated from synchronized MOCAP data. The markers determine the head orientation about a common frame from which the neck inclines. The head inclination is represented by roll-pitch-yaw angles to segment the movement phases. For example, Figure \ref{fig:MOCAP_EEMG_segm} shows the head orientation and EMG data during one trial after the segmentation. \\  
After the segmentation, the raw EMG was filtered with the following steps based on prior work \cite{Diego2022}: the DC component removal, bandpass filter (15-400 Hz), rectification, and local integration by a moving window of 250 ms width. 

\section{RESULTS}  
\subsection{Mechanical characterization}
\begin{figure}[t]
    \begin{center}
     \includegraphics[width=0.98\linewidth]{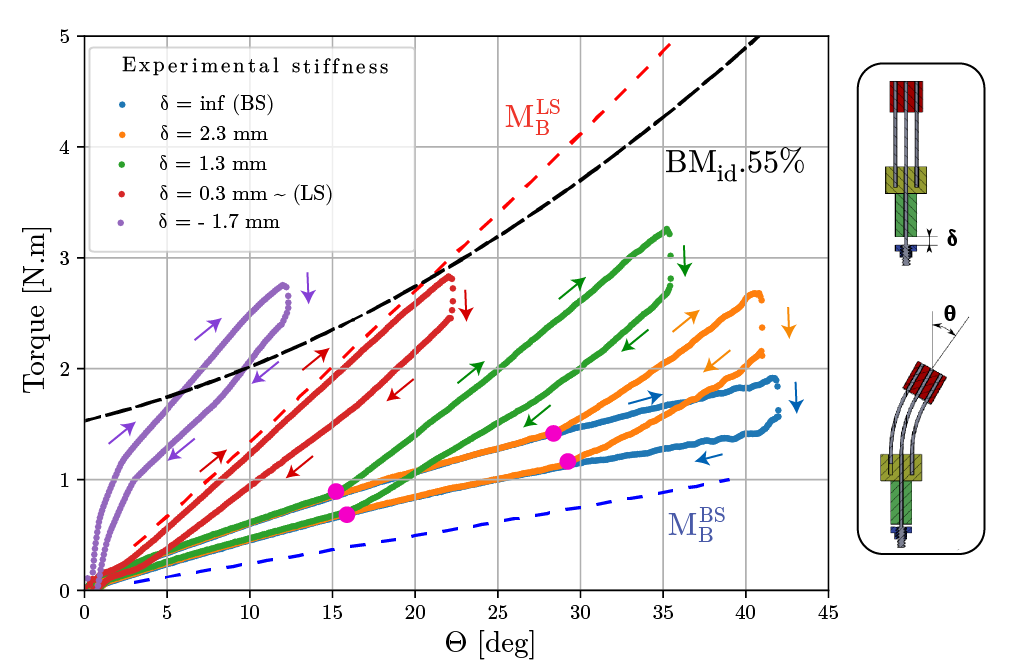}
         \caption{Bending performance. The experiment results present a clear stiffness transition that depends on the parameter $\delta$. The predicted moment  of the actuator model (\ref{Eq:sbar_equation})  for the Loaded and Base Stiffness conditions ($M_{B}^{LS}$ \ref{Eq:highS_I} and $M_{B}^{BS}$ \ref{eq:lowS_I}) show close similarities with experimental results. The device demonstrated to provide up to 55$\%$ of the required Moment $BM_{id}$ deduced from the mocap experiment (\ref{eq:ideal_M}).}
        \label{fig:mech_char_res}
    \end{center}      
    \vspace{-4.2 mm}
\end{figure}

The experimental bending stiffness of the elastic mechanism for different $\delta$ values is contrasted with the theoretical predictions and required assistive moment in Figure \ref{fig:mech_char_res}. As discussed in the previous prototype \cite{Me10122051}, the device presents a hysteresis during the mechanical cycle linked to friction between the sliding parts. The experimental curves show both elastic modes are differentiated where the low-stiffness linear response occurs first followed by a sudden stiffness increase indicated by the magenta dots. This transition occurs earlier as the distance $\delta$ is smaller. By convention, the condition $\delta=\inf$ is denominated Base Stiffness, and when the gap is nearly non-existent, $\delta=0.3mm$ is the Loaded Stiffness, Regarding the theoretical model, the experimental results for the Loaded and Base Stiffness are within the numerical prediction showing agreement between both approaches. Beyond the Loaded Stiffness condition, higher assistive forces can be achieved by compressing the adjustable plate against the rigid restriction ($\delta=-1.7mm$) which preloads the mechanism. In conclusion, these results show that the prototype can compensate up to 55$\%$ of the head weight during sagittal flexion.

\subsection{Human Experiment}
\begin{figure} 
    \centering
  \subfloat[\label{fig:sag_mot} Flexion on the Sagittal Plane]{%
       \includegraphics[width=0.84\linewidth]{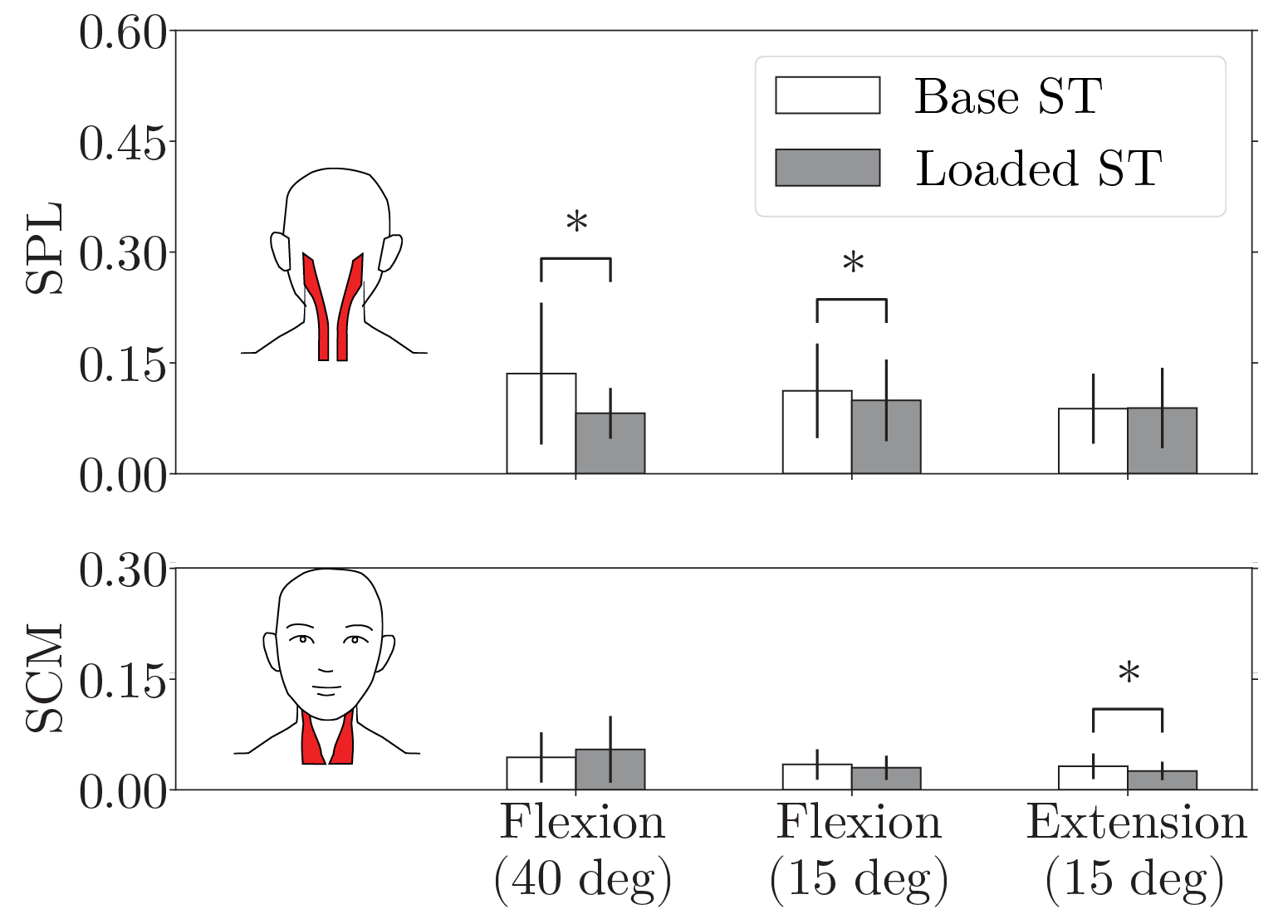}}
    \hfill
  \subfloat[\label{fig:rot_mot} Axial Rotation]{%
        \includegraphics[width=0.84\linewidth]{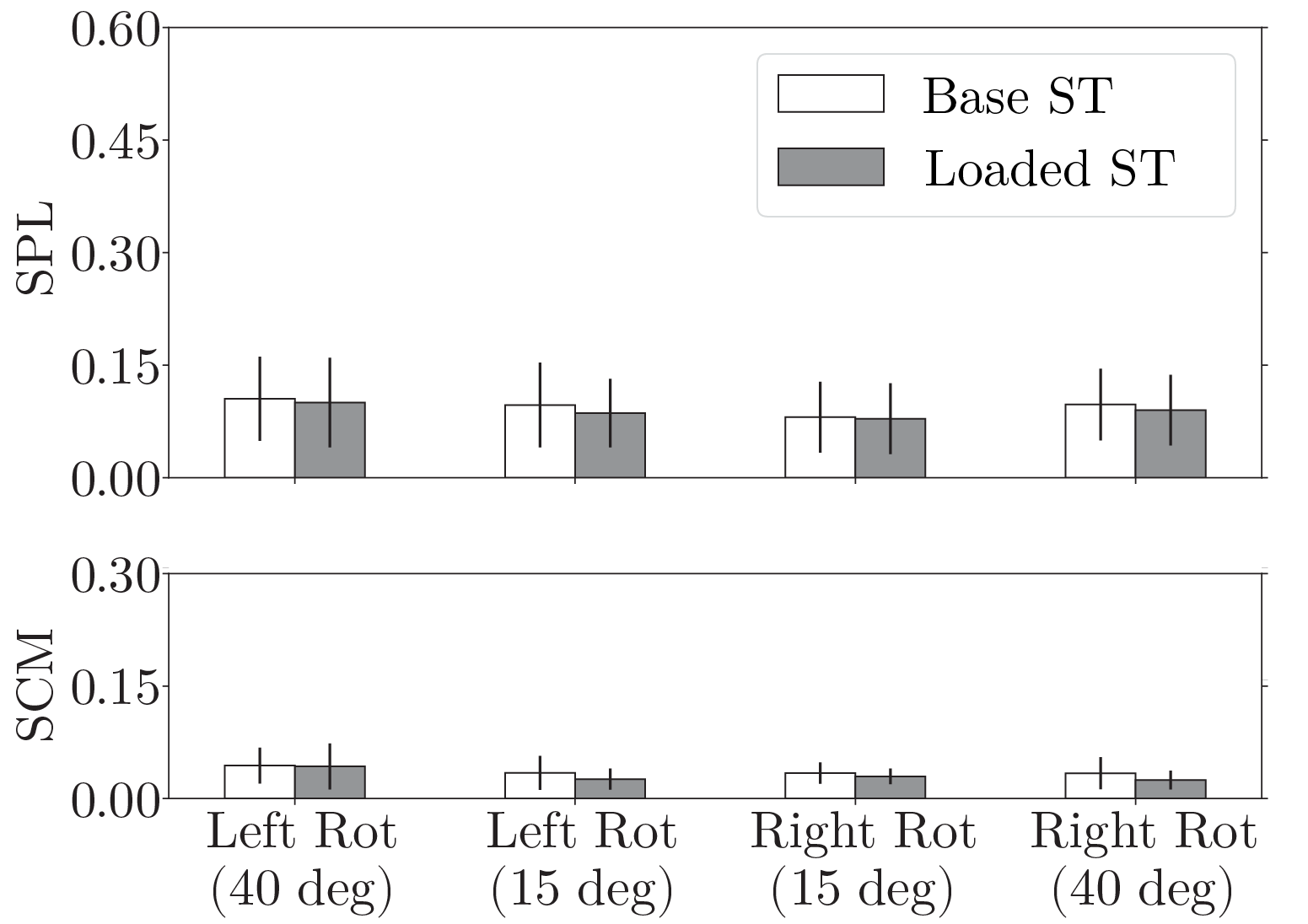}}
  \caption{Comparison of the averaged EMG activity and neck ROM in the Holding Phase between the device's Base and Loaded Stiffness (ST) modes during sagittal flexion and axial rotation. The studied muscles were Sternocleidomastoid (SCM) and Splenius Capitis (SPL). The significant decrement in both SPLs' activity during flexion evidences an increment of the assistive support for the higher stiffness without significantly affecting the effort to perform rotation. The small ROM difference between conditions indicates that the device`s stiffness adjustment does not significantly affect the mobility of the neck}
  \label{fig:EMG_results}
   \vspace{-5mm}
\end{figure}

The average muscle activity for the SCM and SPL group during the Holding Phase is compared between the Loaded and Base Stiffness conditions in Figure \ref{fig:EMG_results} for each neck motion. Each subfigure presents the results for movements on the coronal and transverse plane on a bar plot for each muscle group. Each graph presents the mean value and deviation of the average EMG value from the studied group during the Loaded Stiffness and Base Stiffness conditions, which were compared using the two-sided Wilcoxon Signed Rank Test. Before computing the average EMG, each participant's mean activity during the target motion was calculated from the holding phase defined in the protocol and normalized by the maximum recorded value during the execution of rotational motions, the associated movement to the greater activity of the studied muscles \cite{Yajima2022}.
Additionally, the head inclination was computed to segment the EMG activity and assess the participant`s performance following the GUI instructions. Table \ref{tab:ROM} shows the mean pitch and yaw during the holding phase for flexion and rotation movements respectively:

\begin{table}
  \centering
  \vspace{1.2mm}
  \caption{\label{tab:ROM} Mean angular position during each motion}
  
  \begin{tabular}{c *{4}{S[table-format=-2.0(1), table-align-text-post=false, separate-uncertainty]}}
    \hline
    Motion & \multicolumn{2}{c}{Flexion} & \multicolumn{2}{c}{Rotation} \\
    \cline{2-5}
           & {LS} & {HS} & {LS} & {HS} \\
    \hline  
    40     & 36 \pm 3 & 39 \pm 1 & 30 & 30 \pm 3 \\
    15     & 14 \pm 2 & 14 \pm 1 & 15 & 11 \pm 1 \\
    -15    & -12 \pm 3 & -9 \pm 4 & -15 & -15 \pm 6 \\
           & & & -30 & -31 \pm 3 \\
    \hline
  \end{tabular}
  \vspace{-2mm}
\end{table}

The backward flexion at 40 degrees was omitted because the vertical slider restricted mobility for some cases leading to dispairing results. For similar reasons, lateral flexion results show dissimilar results to the presence of friction in some trials which induced an additional resistance that restricted the mobility on the coronal plane. 

During the backward flexion, the SCM EMG activity significantly decreases while the mean bending angle between both conditions defers 3 degrees within the values uncertainties.

During sagittal flexion, the SPL activity significantly decreases in the Loaded Stiffness condition. Also, there is a slight increase when the bending angle is 40 degrees. 
Finally, there are no major changes during axial rotation regarding the mobility of muscle effort.

\section{DISCUSSION}
The results for flexion on the sagittal plane present evidence that the assistive force of the device can be regulated by adjusting the system stiffness, given that the SPL muscles supporting the head show an EMG activity reduction without significantly affecting mobility. In contrast, the SCM group does not present significant differences in the activity suggesting that the assistive force does not surpass the required value to compensate for the head weight, which can be concluded from the bending test results in Figure \ref{fig:mech_char_res}. This aspect was considered in the mechanical design to prevent over-training the muscles involved in bending the head. Given these results, it is expected that DHS users will experience a similar muscular unloading compensating the muscle weakness and reducing the neck deformity but permitting a wide range of motion when needed. Adjusting the assistive force would allow the user to keep using the weak muscle group potentially preventing or lowering their atrophy.

In contrast with related work, this research presents the first result of a passive device that decreases muscle effort on supporting the head weight during motion on the sagittal plane without overloading the muscles involved in the active flexion. Passive Columbia Neck brace \cite{Zhang20183} actuated by torsional springs shows no statistically significant differences in the average neck muscle activity whether the springs are used or not, while the study on the wheelchair brace developed in Vrije \cite{Mahmood2021} indicates a reduction on the Upper Trapecious at the expense of a significant increment of the SCM group during forward flexion. Studies on the Sheffield snood analyzed motion restriction but no muscle effort \cite{Pancani2016}.  

The EMG results for back flexion are analogous to the previous case. The SCM muscles supporting the head show a muscle activity reduction for the Loaded Stiffness condition while the SPL group is not significantly affected. In conclusion, these results emphasize that the user did not experience major effort for performing this motion, which facilitates executing tasks like reaching objects above the line of sight.

Thirdly, small and large rotations for both sides were analyzed showing a symmetric behavior. Each case shows no significant changes in the muscle activity and mobility between conditions suggesting that the slider guide permits a free rotation regardless of the device`s stiffness.

Regarding the EMG analysis, the results for the approaching and recovery phases were not included due to insignificant changes in muscle activation between conditions. Inter-participant differences in movement speeds affecting muscle activation may occlude a possible EMG reduction. The different velocities were due to the interface`s limited feedback as the reference avatar of the interface instantly moved from the origin to the target position during the motion phases.
A larger scale study with a GUI that guides the user during the three phases could lead to clearer results for the muscle activity during each phase. A field of study still in an early stage of development \cite{Zhang2019}.

A limitation of this study concerns the current prototype`s fitness and compactness. The entire device weighs 2.5 Kg which is way over the recommended value achieved by other authors \cite{Zhang2017}. Given that the elastic module of the system weighs 100 grams, we estimate that the orthosis weight can be reduced to 1.2 Kg by using topology-optimized 3D-printed components. Regarding the fitting, the vest attachment ensures an effective force transfer at the cost of depending on another person to self-fit the device which is a disadvantage. Alternative attachments that facilitate the donning and doffing of the device will be considered in the future.
Another limitation of this work is the absence of a condition where the participant does not wear the device. Including such a condition would permit quantifying the effects of the device`s mechanical imperfections on muscle activity. Despite this additional information that would expand the research scope, we consider that the presented results highlight the device`s capacity to provide adjustable support, which is central for neck orthosis for DHS patients. Future studies will consider these conditions for a more comprehensive device analysis.

In summary, the presented results suggest the proposed device provides adjustable support for the head in the sagittal plane without severely restricting mobility during sagittal flexion or axial rotation. These features could provide an ideal balance of support and mobility to DHS patients that current devices lack. The upcoming experiments will evaluate these aspects.

\section{CONCLUSION}
This work introduces a novel portable passive neck brace for DHS patients that assists head motions in the sagittal plane and permits the free rotation of the neck. The device assistance can be regulated by an elastic mechanism of adjustable stiffness whose actuation principle was theorized and validated through a mechanical test. The neck brace effects on neck muscle activity were studied in 8 healthy participants for different stiffness values of the actuator showing significant decrements in the extensor muscles for motions in the sagittal plane for the higher stiffer conditions. Additionally, no significant changes in mobility or EMG activity were measured during neck rotation suggesting that the device does not restrict such a motion. These results present evidence that the proposed device reduces neck muscle effort while enabling a wide range of motion on the sagittal and axial plane, an essential feature for a life support device for DHS patients. Future evaluations will test this hypothesis on real patients.






\bibliographystyle{IEEEtran}
\bibliography{IEEEabrv,MyCitations}
\end{document}